%% file: neurips_2024.tex
\documentclass{article}

\usepackage[final]{neurips_2024}

\usepackage[utf8]{inputenc} 
\usepackage[T1]{fontenc}    
\usepackage{hyperref}       
\usepackage{url}            
\usepackage{booktabs}       
\usepackage{amsfonts}       
\usepackage{nicefrac}       
\usepackage{microtype}      
\usepackage{xcolor}         
\usepackage{comment}
\usepackage{graphicx}
\usepackage{pdfpages}
\usepackage{subcaption}
\usepackage{tabularx}
\usepackage{microtype}
\usepackage{times}
\usepackage{latexsym}
\usepackage{adjustbox}
\usepackage{multicol}

\usepackage{inconsolata}

\title{Developing Story: Case Studies of Generative AI's Use in Journalism}

\author{
  Natalie Grace Brigham, Chongjiu Gao, Tadayoshi Kohno\\
  \textbf{Franziska Roesner, Niloofar Mireshghallah} \\
  University of Washington \\
  \texttt{\{nbrigham, chongjiu, yoshi, franzi, niloofar\}@cs.washington.edu} \\
}

\begin{document}
\maketitle

\begin{abstract}
\input{latex/abstract}

\end{abstract}

\section{Introduction}
\input{latex/introduction}

\input{latex/relatedWork}

\section{Method}
\input{latex/methodology}

\section{Findings}
\input{latex/results}

\section{Conclusion}
\input{latex/conclusion}

\section{Social Impacts Statement}
\input{latex/socialImpactsStatement}
\section*{Acknowledgements}
\input{latex/acks}

\section*{Limitations}
\input{latex/limitations}

\bibliography{main}
\bibliographystyle{main}

\clearpage
\appendix
\input{latex/appendix}

\end{document}

%% file: latex/abstract.tex
Journalists are among the many users of large language models (LLMs).
To better understand the journalist-AI interactions, 
we conduct a study of LLM usage by two news agencies through browsing the WildChat dataset, identifying candidate interactions, and verifying them by matching to online published articles. 
Our analysis uncovers instances where journalists provide sensitive material such as confidential correspondence with sources or articles from other agencies to the LLM as stimuli and prompt it to generate articles, and publish these machine-generated articles with limited intervention (median output-publication ROUGE-L of 0.62).
Based on our findings, we call for further research into what constitutes responsible use of AI, and the establishment of clear guidelines and best practices on using LLMs 
in a journalistic context.

%% file: latex/introduction.tex
Given the potential of LLMs to assist journalists and increase productivity, several initiatives are aiming to assist news organizations in finding, training, and applying AI-based solutions responsibly~\citep{APAIInitiative, PartnershipOnAIInitiative}.
However, many have raised concerns about the application of LLMs in the field of journalism,
including misinformation~\citep{Pan2023misinfo}, copyright violations~\citep{Karamolegkou2023Copyright}, and privacy implications~\citep{Yao2024Privacy}.
Due to these issues, LLMs can be seen as both a tool for journalism and a threat to journalistic integrity~\citep{Wihbey2024rish}, depending on how they are used.
While previous work surveyed journalists' reported usage of generative AI~\citep{APJournalismReport, Gondwe2023GlobalSouth}, the opaque nature of newsrooms and challenges in detecting AI-generated content~\citep{Chakraborty2023OnTP, Weber-Wulff2023} have left a knowledge gap in the computer science research community regarding its on-the-ground utilization. 
This work aims to bridge this gap by analyzing journalist-AI interactions, including queries, provided materials, intervention levels, and query-to-publish timelines.

To achieve this, we probe the publicly available WildChat dataset~\citep{zhao2024wildchat} of human-chatbot interactions.
We identify potential journalist queries in WildChat, match them to published articles on two news agency websites, and analyze the resulting set of interactions.
Given WildChat's scope and that other agencies likely use generative AI, we refer to the two identified agencies as Agency~A and Agency~B to maintain anonymity (see limitations and ethical considerations).
We categorize the types of tasks for which LLMs were used across Agencies~A and~B and the distribution of input materials employed to generate articles, shown in Figure~\ref{fig:stimuli}.
Inputs include articles from other agencies or private conversations, real case examples of which are depicted in Figures~\ref{fig:case-study-1} and~\ref{fig:case-study-2} (Agencies~B and~A, respectively).
We also analyze the extent of human intervention in article generation by examining the overlap between the model generated drafts and the matched published articles, along with the time between generation and publication.
Finally, we go beyond the WildChat dataset and study machine-generated text on Agency~A's and~B's websites using GPTZero~\citep{tian2023gptzero}, a state-of-the-art commercial LLM-generation detector.

Our findings offer insights about the role of LLMs in journalism and potential impacts to journalistic integrity, as we find substantial overlap between the model outputs and the published articles with a median ROUGE-L~\citep{lin-2004-rouge} score of 0.62, and a prompt-to-publication span of a single day.
Indeed, $18\%$ of the identified input stimuli are other agencies' news articles and $9\%$ are potential private conversations, risking privacy breaches by sharing the data with the LLM-provider and publicly via WildChat.
Our study points to a need for better AI literacy and education for model practitioners.
A potential path forward could draw from human-computer interaction and usable security and privacy research on ``nudging'' users toward beneficial behaviors~\citep{Acquisti2017nudge}, which may offer a promising solution in this scenario as well. 
We also encourage the journalistic and computer science communities to iteratively refine guidelines~\citep{RobotJournalismGuidelines, NYTGuidelines, CombinedGuidelines} for the use of AI in journalism as technologies like LLMs become more advanced and ubiquitous.

%% file: latex/relatedWork.tex
\noindent\textbf{Related work.} Our work relates to studies on human-chatbot interactions for assistive writing, such as in academic paper writing \citep{Liang2024Papers} and in the peer-review process~\citep{Liang2024Reviews}, but differs in topic and methodology, providing visibility into interactions and ground truth for machine-generated articles. We provide an extended study of related work in Appendix~\ref{app:sec:related}.

%% file: latex/methodology.tex
\noindent\textbf{WildChat.}
Since our aim is to closely study the type of queries made by journalists to LLMs and to uncover interventions they make to model outputs before publication, we probe 
WildChat~\citep{zhao2024wildchat}, a publicly available dataset of 650k\footnote{Recently a 1M version of this dataset was released, however our study is conducted on an earlier version.} conversations collected by offering free access to GPT-3.5 and GPT-4 to users, for potential candidate interactions.

\begin{figure*}[t]
  \centering
\includegraphics[width=\textwidth]{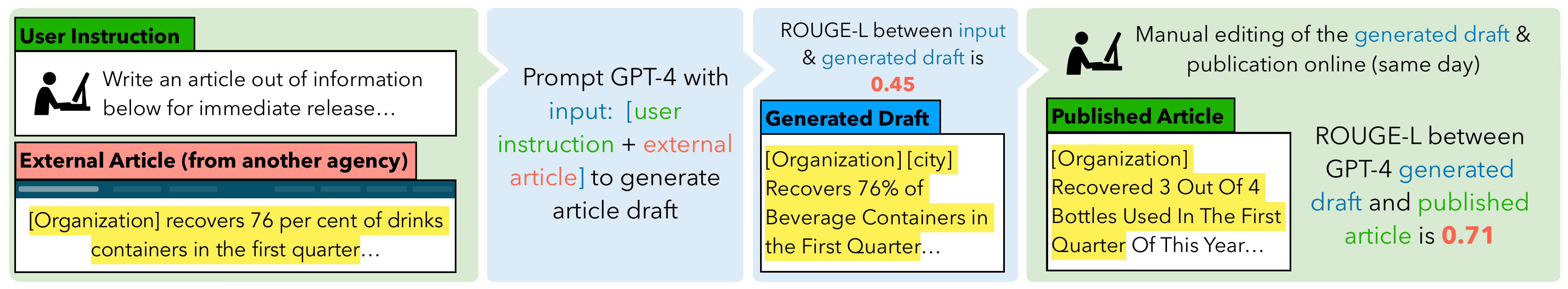}
  \caption{Case study of a single-turn journalist-LLM interaction for article generation where an external article by another agency is used as the input stimulus. The generated draft is published by the journalist with little modification as the ROUGE-L between the published article we identified and the generation is 0.71. The~`[' and~`]' symbols denote portions of the text that have been replaced to minimize identifiability.}
  \label{fig:case-study-1}
\end{figure*}

\begin{table*}[]
  \begin{minipage}[c]{0.42\linewidth}
    \centering
    \begin{tabular}{p{75pt}lll} 
    \toprule 
    \textbf{Agency} & \textbf{A} & \textbf{B} & \textbf{Sum}\\
    \midrule
    Conversations & 62 & 16 & 78 \\
    Turns & 107 & 41  & 148 \\
    Verified published articles & 79 & 32 & 111\\
    \bottomrule
    \end{tabular}
    \caption{Evidence of AI-assisted journalism from WildChat}
    \label{tab:summary}
  \end{minipage}
  \hfill
  \begin{minipage}[c]{0.5\linewidth}
    \centering
    \begin{tabular}{lll}
    \toprule
    \textbf{Task (turn count)} & \textbf{Agency A} & \textbf{Agency B}\\
    \midrule
    Article generation & 89 & 34 \\
    Headline generation & 18 & 1 \\
    Article editing & 0 & 6 \\
    \bottomrule
    \end{tabular}
    \caption{Turn counts for each task type across both agencies.}
    \label{tab:task-type}
  \end{minipage}
  \vspace{-1ex}
\end{table*}

\noindent\textbf{Identifying  journalist queries.}
We initially identified conversations with $4$ or more PII types using an NER model\footnote{\texttt{lakshyakh93/deberta\_finetuned\_pii}}, as we were interested in identifying sensitive disclosures. This yielded a $5k$ turn subset of the data. We then manually inspected this set and  found instances of article generation for two different news agencies: Agency~A, a local news outlet based in southern California, and Agency~B, a local news platform covering a small nation in the Mediterranean region.
Using location-based keywords for these agencies, we filtered the entire 650k conversation WildChat dataset and manually reviewed the results to find additional conversations related to these agencies. 

We acknowledge that this process may result in false negatives, missing cases of journalistic activity in WildChat that we did not identify.
(See Section~\ref{sec:social-impacts} for reasons why we upload conversation IDs but do not mention the specific agencies in the main text.)
To match the output of conversational turns to published articles and verify that the queries were made by the journalists, we searched the identified agencies' websites for articles with highly similar content to the generated text. 

\noindent\textbf{Categorizing tasks.} For each verified conversational turn, based on the user's prompt, we classify the journalistic tasks:
(1)~\textit{article generation} which involves the LLM creating a new article from the provided user instruction and input material, 
(2) \textit{headline generation} which involves generating a headline for a given article, and (3)
\textit{article editing} which involves the user requesting edits to a provided draft article.

\noindent\textbf{Categorizing input stimuli.} By stimulus, we refer to the input material provided to the LLM assistant as source or context.
One researcher examined all user prompts for \textit{article generation}, developed a codebook that was reviewed by the team, and, to ensure coding consistency and retain interpretive nuance ~\citep{Qual2, Qual3}, coded all stimuli.
Then, a second researcher independently reviewed and validated the coding~\citep{Qual1}. 
Stimuli was only coded as an externally sourced type (e.g., press release, external news article) when verbatim text matched online sources.
Thus, these codes represent a lower bound on the external source material. 

\noindent\textbf{Measuring journalist intervention.} As a proxy for how much the journalists modify the model's output before publication online, we report the ROUGE-L recall score~\citep{lin-2004-rouge} between the generated output (used as source) and the matched online article. This score reflects the longest common subsequence between the two text sequences, normalized to the length of the source. We also report this score over the model's input and output.

\noindent\textbf{Detecting LLM-generated articles beyond WildChat.} To extend our study beyond the WildChat dataset, we scrape articles from the two identified news agencies' websites from between January 1, 2020 and April 15, 2024, use GPTZero~\citep{tian2023gptzero}, a state-of-the-art commercial machine-generated text detection method to study the prevalence of LLM use, and report our findings.

%% file: latex/results.tex
\begin{figure*}[t]
  \centering
    \includegraphics[width=\textwidth]{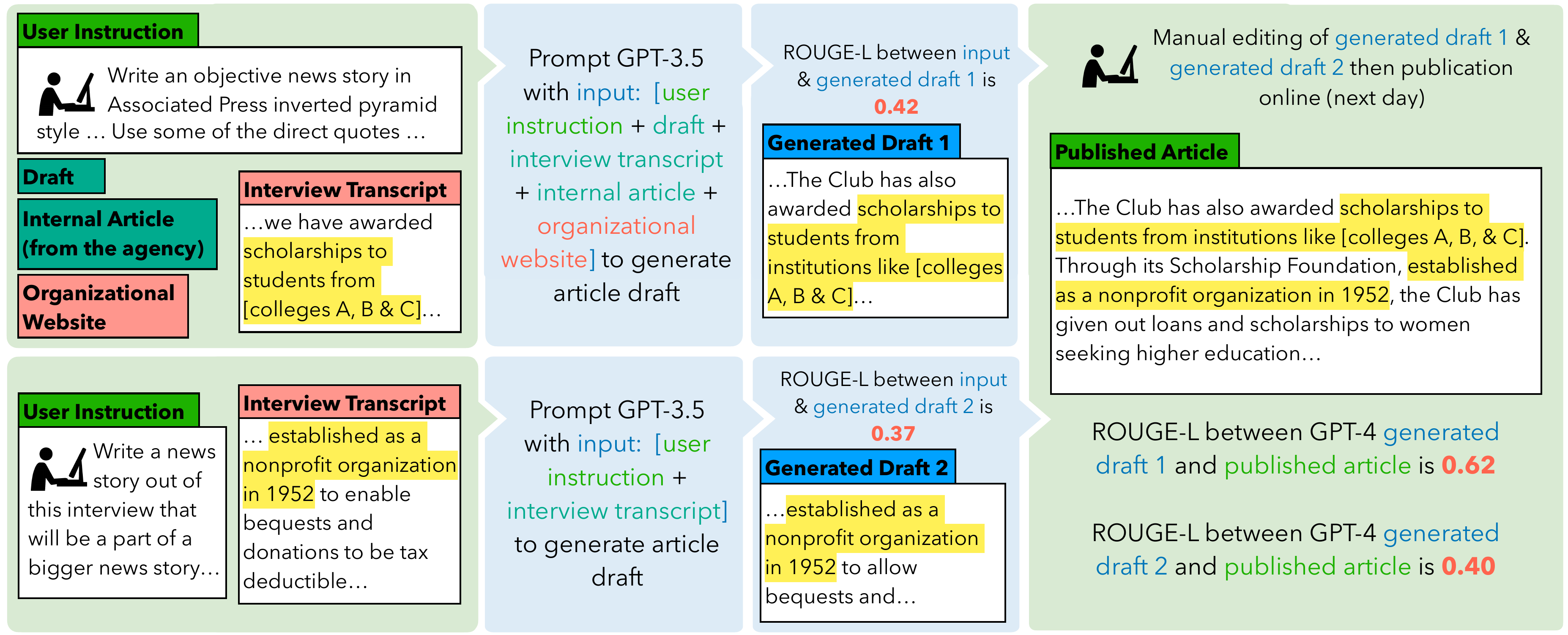}
  \caption{Case study of a multi-turn article generation using multiple stimuli, including an internal article from the same agency and an interview transcript. The~`[' and~`]' symbols denote portions of the text that have been replaced to minimize identifiability.}
  \label{fig:case-study-2}
  \vspace{-3ex}
\end{figure*}

\begin{table*}[t]
  \centering
  \begin{tabularx}{\textwidth}{Xll}
    \toprule
    \textbf{Stimuli type} & \textbf{Agency A} & \textbf{Agency B} \\
    \toprule
    \textbf{Draft or other}: Draft of article or unidentified material & 39 & 9 \\
    \midrule
    \textbf{Press release}: Article on the same topic released by the subject of the story & 30 & 4 \\
    \midrule
    \textbf{External news article}: Article on the same topic from another news agency & 18 & 15 \\
    \midrule
    \textbf{Interview}: Transcript or written interview responses & 14 & 2 \\
    \midrule
    \textbf{Organizational report}: Official report from a city or country government or organization (e.g., a school district's 100 Day Plan) & 12 & 2 \\
    \midrule
    \textbf{Event posting}: Posting of event details (e.g., Meetup, event website) & 12 & 0 \\
    \midrule
    \textbf{Organization website}: General information on an organization, company, or person (e.g., ``About'' page) & 9 & 0 \\
    \midrule
    \textbf{Intra-agency news article}: Article on the same topic from within the agency & 1 & 8 \\
    \midrule
    \textbf{Email}: Message from a source or editor about a story idea & 0 & 5 \\
    \midrule
    \textbf{Social media post}: Post on social media (e.g., LinkedIn, GoFundMe) & 2 & 0 \\
    \bottomrule
  \end{tabularx}
  \caption{Types of stimulus used to generate articles in the identified WildChat activity and their frequencies across Agencies~A and~B, sorted in decreasing order from top to bottom by their combined frequencies.}
  \label{tab:stimuli}
  \vspace{-3ex}
\end{table*}

Table~\ref{tab:summary} summarizes the identified content from WildChat, with the last row showing the number of online articles we could match to the identified activity.
We select two case studies that represent archetypal examples of different article generation behaviors, and use them for demonstrations throughout this section:
Figure~\ref{fig:case-study-1}, a case of a single-turn article generation (from Agency~B) and
Figure~\ref{fig:case-study-2}, a multi-turn process of article generation (from Agency~A). 
Below we discuss our observations from the two agencies. 

\begin{figure*}[t]
  \begin{minipage}[c]{0.5\linewidth}
    \centering
    \includegraphics[width=\linewidth]{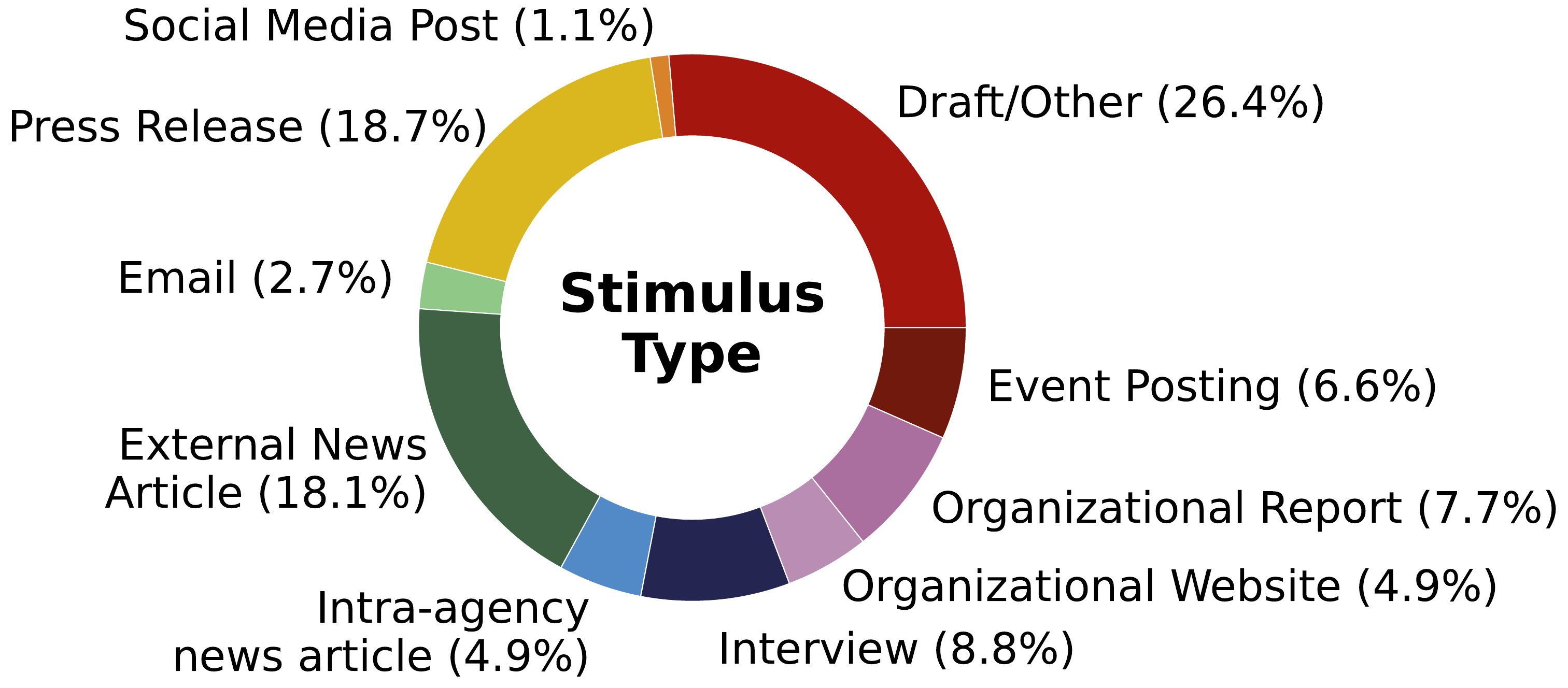}
    \vspace{1ex}
    \caption{The distribution of different input stimulus material types provided by journalists to LLMs, over the WildChat conversations that are matched to online articles from the two identified news agencies.}
    \vspace{-3ex}
    \label{fig:stimuli}
  \end{minipage}
  \hfill
  \begin{minipage}[c]{0.47\linewidth}
  \centering
    \includegraphics[width=\linewidth]{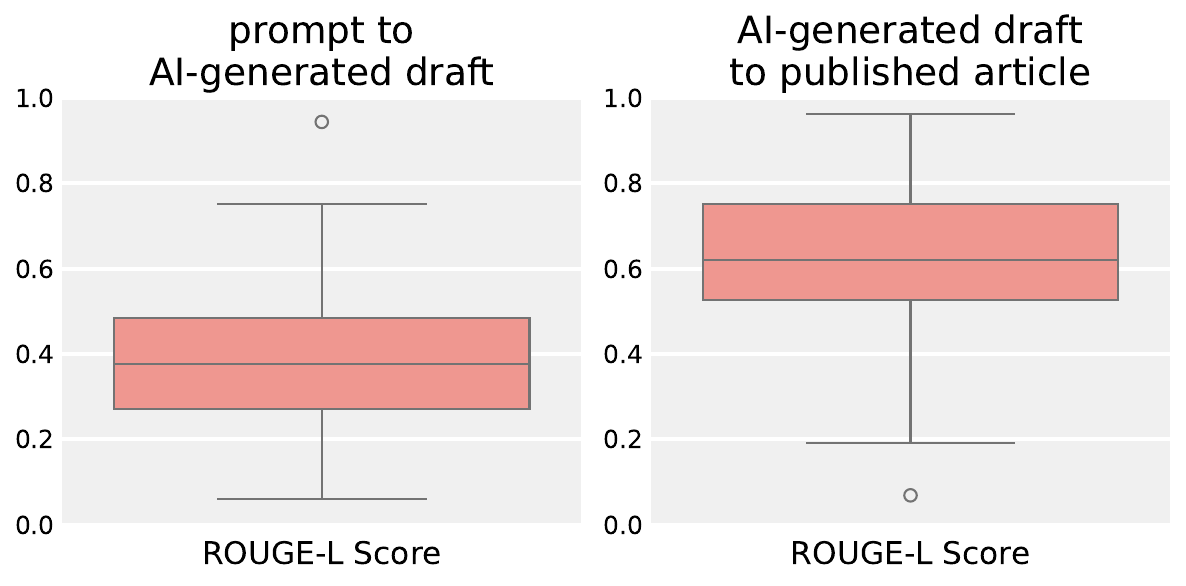}
    \caption{Box plots of ROUGE-L scores between outputs from users' article generations prompts and LLM response (left) as well as the LLM-generated article and the corresponding published article (right).}
  \label{fig:rouge}
  \end{minipage}
\end{figure*}

\noindent\textbf{What do journalists in our study prompt LLMs with?}
\label{sec:stimuli}
Table~\ref{tab:stimuli} provides input stimuli types, descriptions, and frequencies, Figure~\ref{fig:stimuli} shows the distribution of input stimulus types, and Appendix~\ref{app:sec:stimuli} provides additional results for this section.
Across the 148 turns, there is a total of 182 stimuli used (all turns had at least one stimulus as input and some had multiple, some turns had multiple instance of the same stimulus type).
Stimuli from external sources (i.e., news articles from other agencies, press releases, organizational reports, event postings, organizations' websites, social media posts) accounted for over two-thirds (83/137) of the identified stimuli for Agency~A and about half (21/45) for Agency~B. 
We also observe internally sourced stimuli types, such as draft material and articles originating from within the agency.

In Figure~\ref{fig:case-study-1}, we observe Agency~B using a news article from another agency as a stimulus, a practice we identified relatively frequently for both agencies. 
In another case of this, a user from Agency~B specified,
\textit{``write a new article out of the information in this article, do not make it obvious you are taking information from them but in very sensitive information give them credit.''}
As academics, where attribution is key, we found the creation of new content from third-party content, which may be copyrighted, as well as intentionally obscuring such usage, concerning.

Both agencies incorporated interviews into their input prompts. These interviews were presented as written logs from emails or chat applications, or as transcriptions of recorded conversations. While most interviews did not contain sensitive information, one instance included a complete WhatsApp conversation that disclosed a source's phone number and sensitive details about their children, including health conditions. 
This raises concerns about exposing such data to the LLM and through WildChat, without the source's consent or awareness.

In terms of the tasks users specified in their inputs, the vast majority of turns, 83.1\%, are for article generation, 14.5\% are headline generation, and a minority are article editing.
We depict the distribution of task types in Appendix~\ref{app:sec:task-type}.

\noindent\textbf{How much do journalists in our study modify model outputs before publication?} 
\label{sec:rouge}
Figure~\ref{fig:rouge} and Appendix~\ref{app:sec:rouge} depict ROUGE-L distributions for prompts to GPT outputs and those outputs to published articles across all identified activity and for each agency, respectively.
The median ROUGE-L score between the machine-generated drafts and published article text is 0.62 which indicates limited human editing before publication.
As a point of reference, a ROUGE score of 0.5 is considered high in privacy and policy domains~\citep{huang2023privacy,maclaughlin2020source}.
Figure~\ref{fig:case-study-1} depicts a specific use where the ROUGE-L score between the machine-generated draft and the published article is high at 0.71, indicating high overlap with the source text. 
In Figure~\ref{fig:case-study-2}, the ROUGE-L scores between the drafts and published articles are slightly lower but still relatively high. We also found a separate query (see Appendix~\ref{app:sec:case-study}) to generate the headline for this article, and the user publishes the machine-generated headline with no editing. 

\begin{figure}[t]
  \centering
    \includegraphics[width=0.7\textwidth]{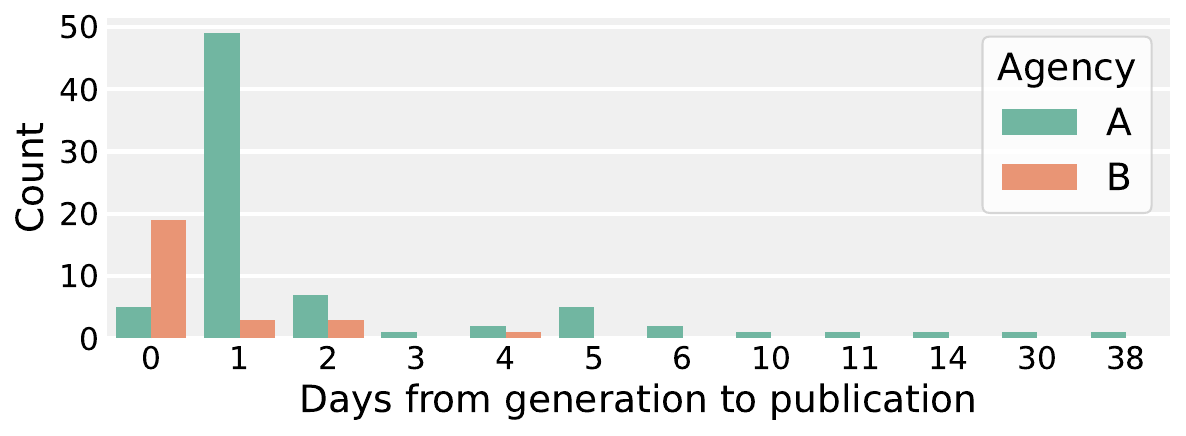}
  \caption{Distribution of days from generation (based on date in WildChat) to publication (based on published article's date) for published articles matched to turns of article generation in WildChat.}
  \label{fig:days-to-publication}
  \vspace{-2ex}
\end{figure}

\noindent\textbf{What is the prompt to publication time?}
For Agency~A, the majority (48/79) of verified articles were published one day after the recorded activity in WildChat. Similarly, the majority (19/32) of verified articles for Agency B were published on the same day as the recorded activity. 
The days of human editing time before publication seen in Figures~\ref{fig:case-study-1} and \ref{fig:case-study-2} are consistent with these findings, as illustrated in Figure~\ref{fig:days-to-publication}.

\noindent\textbf{Are there more LLM-generated articles?}
\label{sec:GPTZero}
To study the temporal trends in using LLMs for writing articles, we scrape articles from Agencies~A and B, between January 1, 2020, and April 15, 2024, subsample 585 articles from each agency uniformly and utilize GPTZero to detect which articles might be machine-generated.
Our findings, depicted in Figure~\ref{fig:gptzero}, suggest that there are likely many additional published articles that may have been generated using ChatGPT. 
Furthermore, we observe a noticeable increase in articles with higher probabilities of machine generations  following ChatGPT's release in November 2022.

%% file: latex/conclusion.tex
We investigate the use of commercial LLMs in journalism by analyzing conversations from the WildChat dataset and matching them to published articles online. 
Our findings reveal the use of potentially unethical material to generate articles, limited human oversight on model outputs before publication, and the use of LLMs by the identified agencies beyond the scope of WildChat. 
These results suggest continuous, increasing generative AI use for news generation and can serve as additional motivation for the co-evolution of guidelines for responsible AI journalism in collaboration between the computer science and journalism communities.

%% file: latex/socialImpactsStatement.tex
\begin{figure*}[t]
    \centering
    \begin{subfigure}{0.48\linewidth}
    \centering  \includegraphics[width=0.98\linewidth]{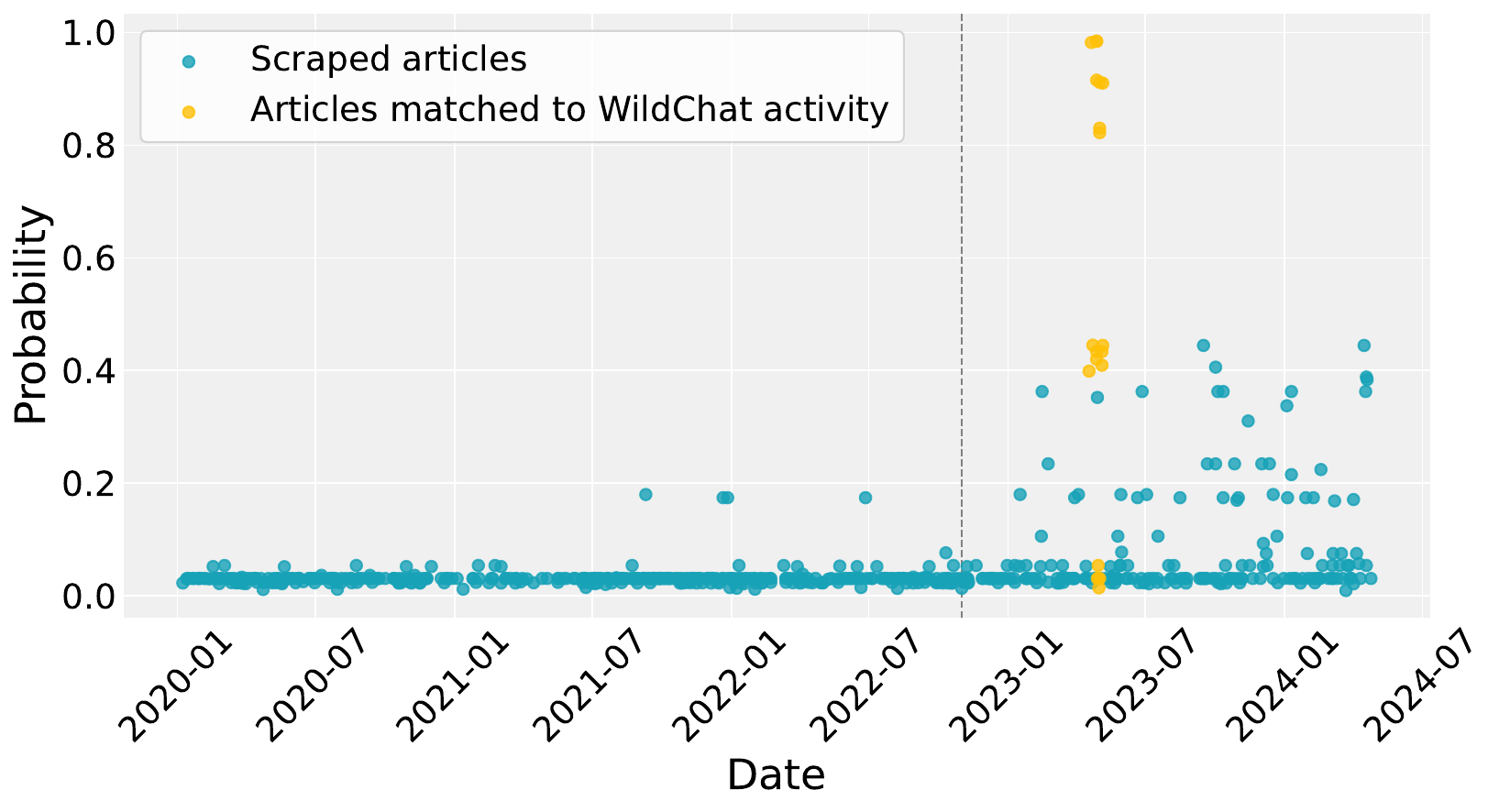}
     \footnotesize
     \caption{Agency A}
    \end{subfigure}
    \begin{subfigure}{0.48\linewidth}
    \centering
\includegraphics[width=\linewidth]{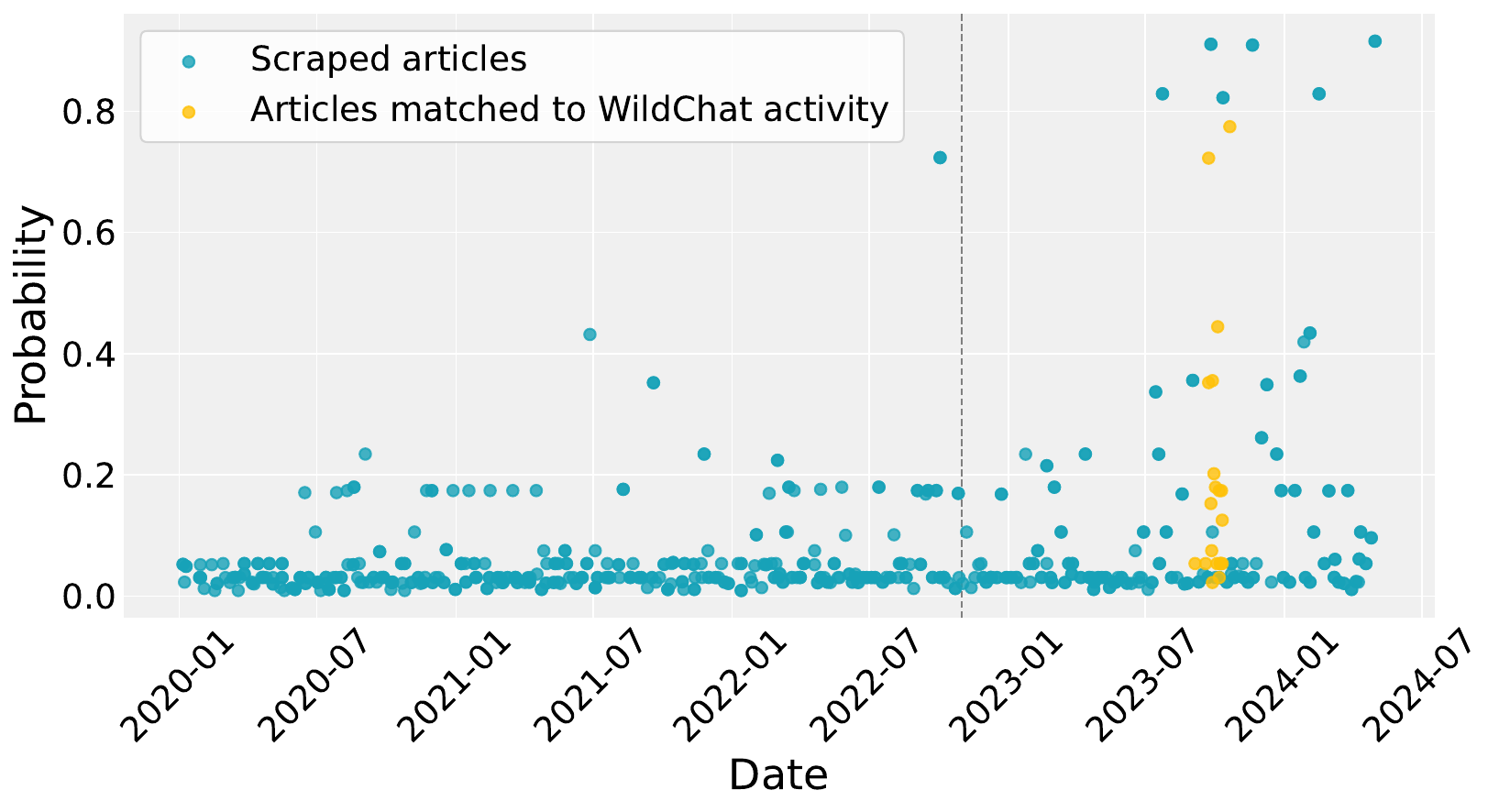}
    \footnotesize 
     \caption{Agency B}
    \end{subfigure}
  \caption{GPTZero probability scores for randomly selected articles from our scrape as well as the published articles matched to WildChat. The grey vertical line indicates November 2022, when ChatGPT was released.} 
     \label{fig:gptzero}
     \vspace{-2ex}
\end{figure*}

\label{sec:social-impacts}

This study is motivated by the recognition that generative AI capabilities are rapidly evolving and their potential impacts on journalism, both beneficial and detrimental, call for thoughtful and thorough research.
Our goal is to provide insights that can inform guidelines and strategies for the responsible use of AI in journalism.
We explicitly do not seek to speak to how \emph{all} news agencies use or might use LLMs. Rather, through these case studies, we seek to develop an informed understanding of how \emph{some} news agencies have used LLMs. Whether their practices are unique to them, or widespread, an understanding of their practices can contribute to a broader discussion about current and potential future uses of LLMs in journalism.

\noindent\textbf{Positionality.}
This work stems from a collaboration between researchers in the NLP and computer security research communities. Consequently, our focus is on the potential risks and harms associated with the use of LLMs in journalism. We acknowledge that there are potential benefits of LLMs in journalism that other perspectives and studies may highlight.

\noindent\textbf{Anonymization.}
We hypothesize that Agencies~A and~B are not unique cases of news organizations using generative AI, but rather examples of a broader phenomenon that happen to be present in LLM system usage.
Furthermore, the intent of this work is not to ``call out'' specific agencies or individual users, but to investigate and better understand the actual use of generative AI in journalism. 
Accordingly, we have anonymized the identities of the agencies in this paper.

Although we anonymize the agencies' names and censor identifying information, we provide metadata and verbatim text that could be used to identify them. 
We include this information for scientific rigor and thoroughness.
However, we do not encourage or condone the use of this data beyond our stated purpose of understanding AI usage in journalism. 
Our intent is to advance knowledge, not target or harm any individuals or organizations. 

\noindent\textbf{Mitigation of privacy concerns.}
Given the sensitive nature of the conversations identified, we notified the WildChat dataset authors of our findings. In response, they removed the identified conversations from the publicly available version of the dataset.

%% file: latex/acks.tex
This work was supported in part by NSF Award \#2205171. We thank Yuntian Deng, Yanai Elazar, Susan McGregor, Stefan Milne, Melanie Sclar, and Maria Antoniak for insightful discussions and feedback.

%% file: latex/limitations.tex
We acknowledge three main limitations in our study: (1) We are constrained by the WildChat dataset itself, which limits our ability to generalize results beyond the two identified agencies and only allows examination of interactions with GPT-4 and GPT-3.5. (2) There may be interactions we were unable to identify within the dataset, though we conducted additional searches and manual reviews to mitigate this. (3) Concerns exist regarding the effectiveness of detecting machine-generated text~\citep{Solaiman2019ReleaseSA, Chakraborty2023OnTP, Sadasivan2023CanAT, Weber-Wulff2023}, and we recognize that GPTZero, the tool used for detection in this study, does not provide ground truth predictions.

%% file: latex/appendix.tex
\section{Extended Related Work}
\label{app:sec:related}
Research at the intersection of journalism and generative AI is still in its early stages. However, work has begun to investigate its potential and actual uses in the field. 
As part of the Associated Press's Local News AI Initiative,~\citeauthor{APJournalismReport}\ surveyed 292 individuals in the news industry about their use and opinions of generative AI in newsrooms. The survey revealed that generative AI is already being used for tasks, such as content production, and changing workflows and role definitions in the newsroom.
Gondwe~\citep{Gondwe2023GlobalSouth} investigated the use of ChatGPT by journalists in sub-Saharan Africa, finding that the system's training on a non-representative African corpus limits its utility for the studied population.

Bdoor and Habes~\citep{Bdoor2024Newsroom} experimented with using GPT to generate news content and discussed the trade-offs around its adoption in the newsroom. Pavlik~\citep{Pavlik2023Collaborating} `co-authored' an essay with ChatGPT to demonstrate both the capacity and limitations of generative AI in journalism and media education.

To address concerns about AI-generated news media, Kumarage et al.~\citep{kumarage2023jguard} developed J-Guard, a framework for directing supervised AI-generated text detectors to identify AI-generated news articles.

\section{Task Type}
\begin{figure}[tb]
    \centering
    \begin{subfigure}{0.32\linewidth}
    \centering
\includegraphics[width=0.95\linewidth]{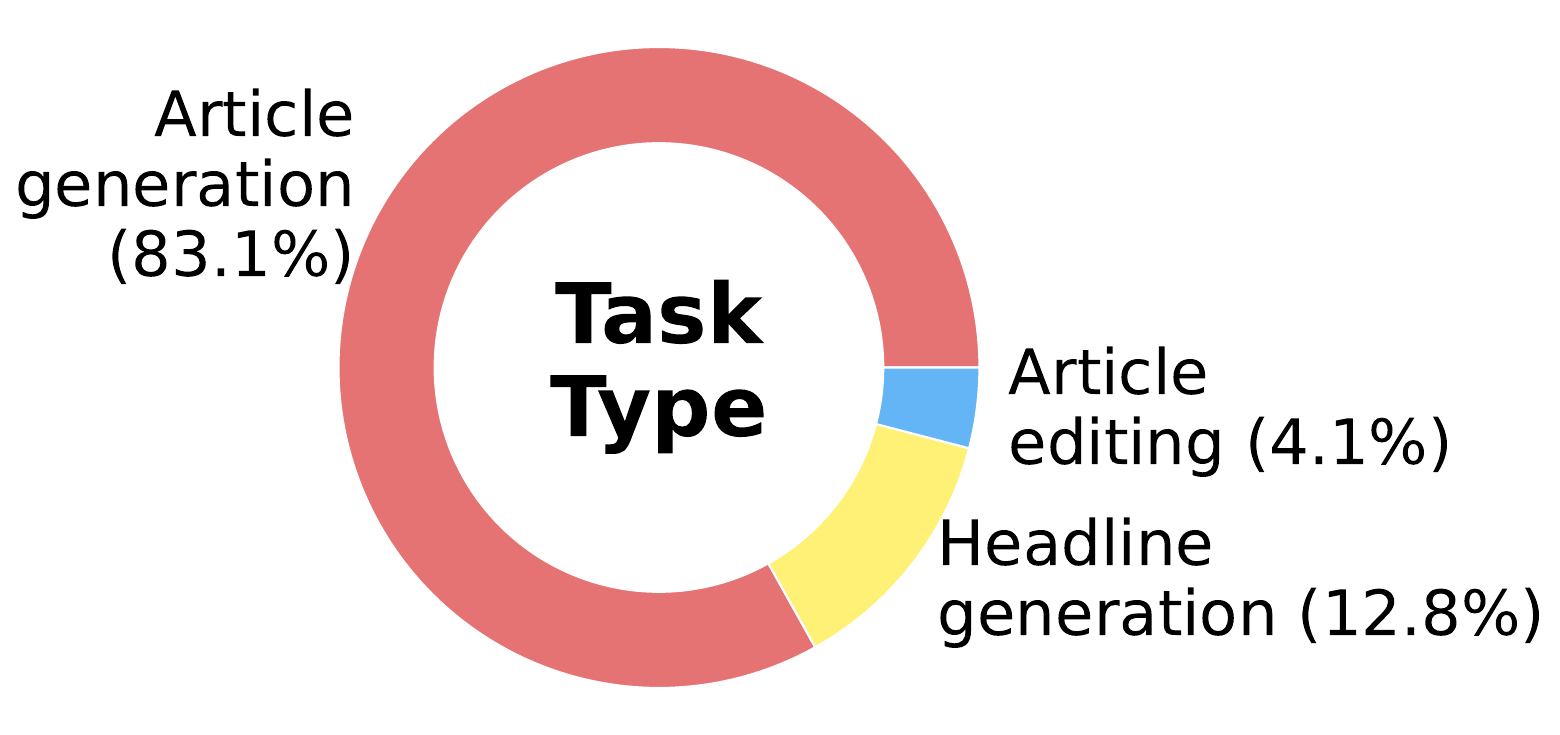}
     \footnotesize
     \caption{Aggregated over both agencies}
    \end{subfigure}
    \begin{subfigure}{0.32\linewidth}
    \centering
\includegraphics[width=0.95\linewidth]{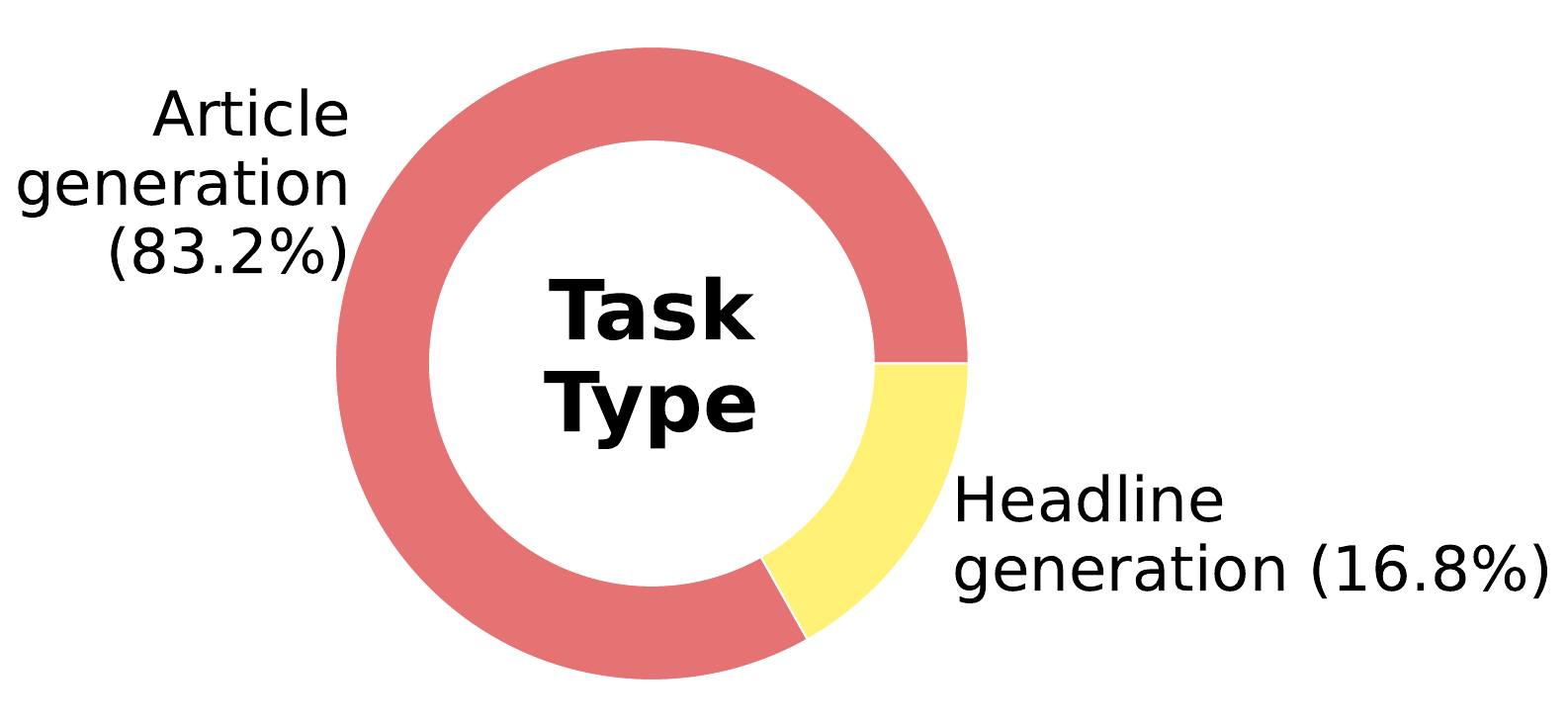}
     \footnotesize
     \caption{Agency A}
    \end{subfigure}
    \begin{subfigure}{0.32\linewidth}
    \centering
\includegraphics[width=0.95\linewidth]{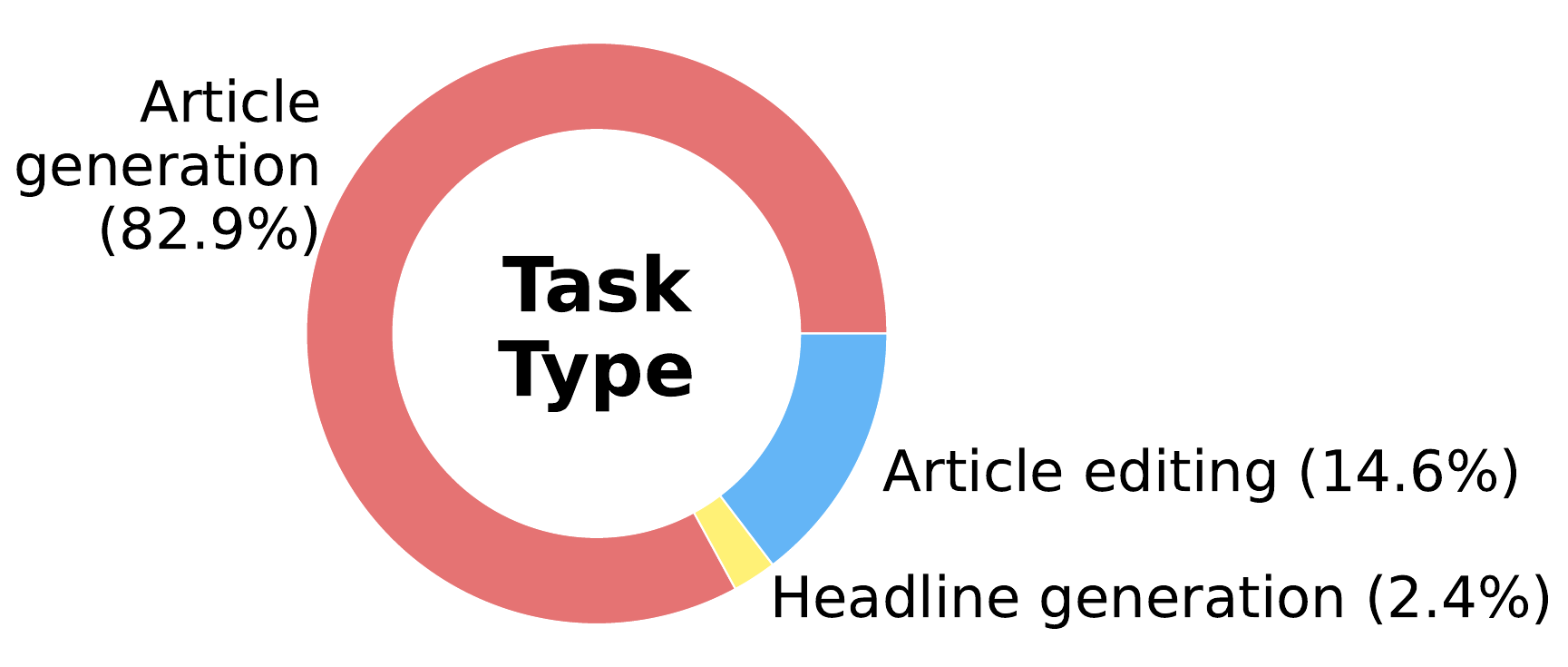}
    \footnotesize 
     \caption{Agency B}
    \end{subfigure}
  \caption{The distribution of different task types across turns from verified Wildchat journalist-LLM conversations.} 
    \label{fig:task-type-app}
\end{figure}

Figure~\ref{fig:task-type-app} provide breakdown of task types.
\label{app:sec:task-type}

\section{Additional Case Study}\label{app:sec:case-study}
\begin{figure*}[tb]
  \centering
    \includegraphics[width=\textwidth]{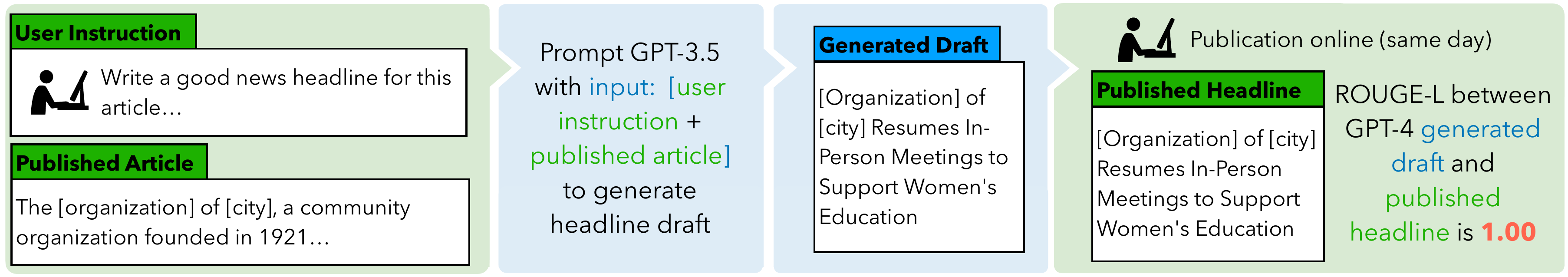}
  \caption{A case study of headline generation, extending from Figure~\ref{fig:case-study-2}. The~`[' and~`]' symbols denote portions of the text that have been replaced to minimize identifiability.}
  \label{fig:additional-case-study}
\end{figure*}

Figure~\ref{fig:additional-case-study} is an extension of Figure~\ref{fig:case-study-2} where the user generates a headline for the machine-drafted article.

\section{Stimuli}
\label{app:sec:stimuli}
\begin{figure}[]
    \centering
    \begin{subfigure}{0.49\linewidth}
    \centering
\includegraphics[width=0.95\linewidth]{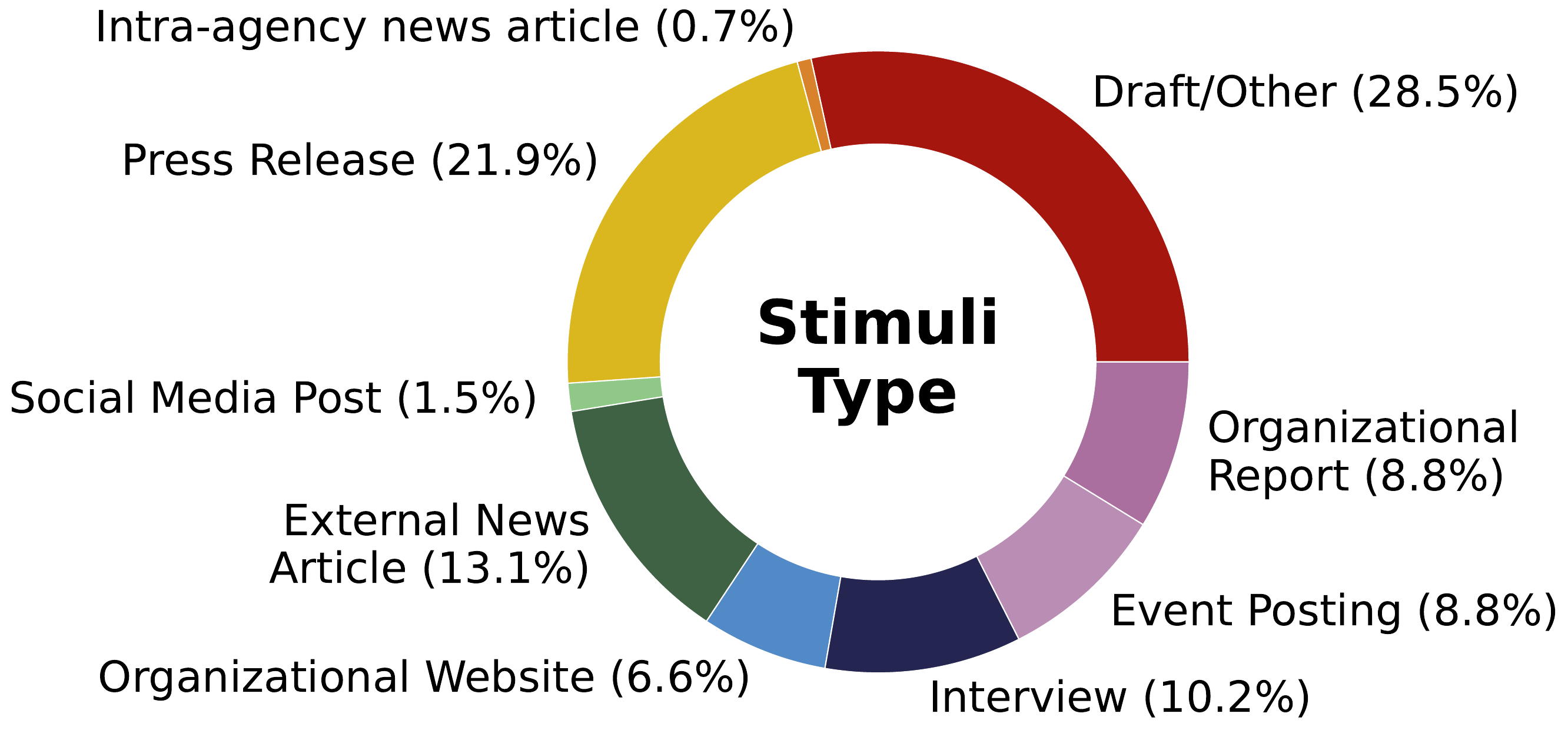}
     \footnotesize
     \caption{Agency A}
    \end{subfigure}
    \begin{subfigure}{0.49\linewidth}
    \centering
\includegraphics[width=0.95\linewidth]{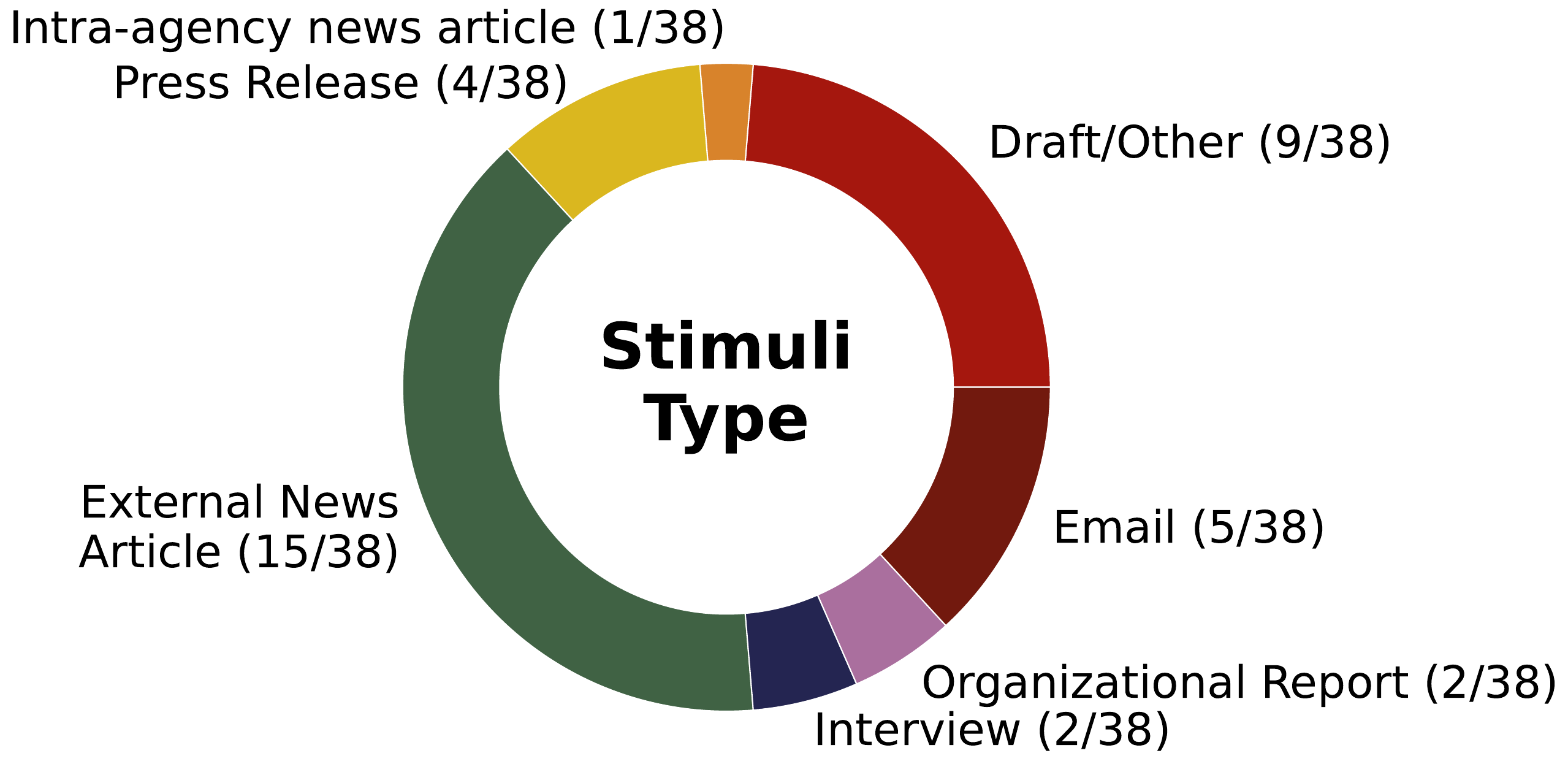}
    \footnotesize 
     \caption{Agency B}
      \label{fig:stimuli-A}
    \end{subfigure}
  \caption{The distribution of input stimuli types over the verified Wildchat journalist-LLM interactions for both agencies.} 
    \label{fig:stimuli-app}
\end{figure}

Table~\ref{tab:freq-stimuli} includes the top five most frequent combinations of stimuli used in a turn of article generation. Figure~\ref{fig:stimuli-app} visualize the distributions of stimuli types for Agencies~A and~B, respectively.

\begin{table*}[]
  \centering
  \begin{tabular}{lp{40pt}lp{40pt}}
    \toprule
    Stimuli type(s) & Agency A frequency & Stimuli type(s) & Agency B frequency\\
    \cmidrule(lr){1-2} \cmidrule(lr){3-4}
    Draft or other & 16 & External news article & 13 \\
    \cmidrule(lr){1-2} \cmidrule(lr){3-4}
    Press release & 16 & Draft or other & 5 \\
    \cmidrule(lr){1-2} \cmidrule(lr){3-4}
    Organizational report & 11 & Press release & 3 \\
    \cmidrule(lr){1-2} \cmidrule(lr){3-4}
    External news article & 6 & Email & 3 \\
    \cmidrule(lr){1-2} \cmidrule(lr){3-4}
    Draft or other, external news article & 5 & Organizational report & 2 \\
    \bottomrule
  \end{tabular}
  \caption{Top five most frequent combinations of stimuli types used in an individual turn of article generation.}
  \label{tab:freq-stimuli}
\end{table*}

\begin{figure}[]
    \centering
    \begin{subfigure}{0.5\linewidth}
    \centering
     \includegraphics[width=0.95\linewidth]{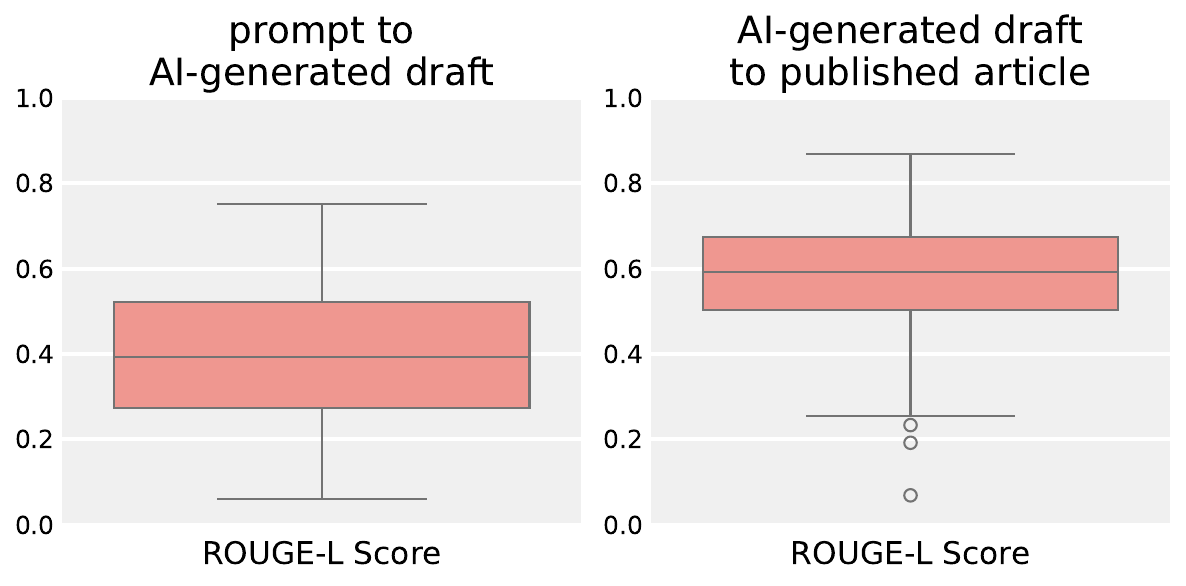}
     \footnotesize
     \caption{Agency A}
     \label{fig:rouge-A}
    \end{subfigure}%
    \begin{subfigure}{0.5\linewidth}
    \centering
     \includegraphics[width=0.95\linewidth]{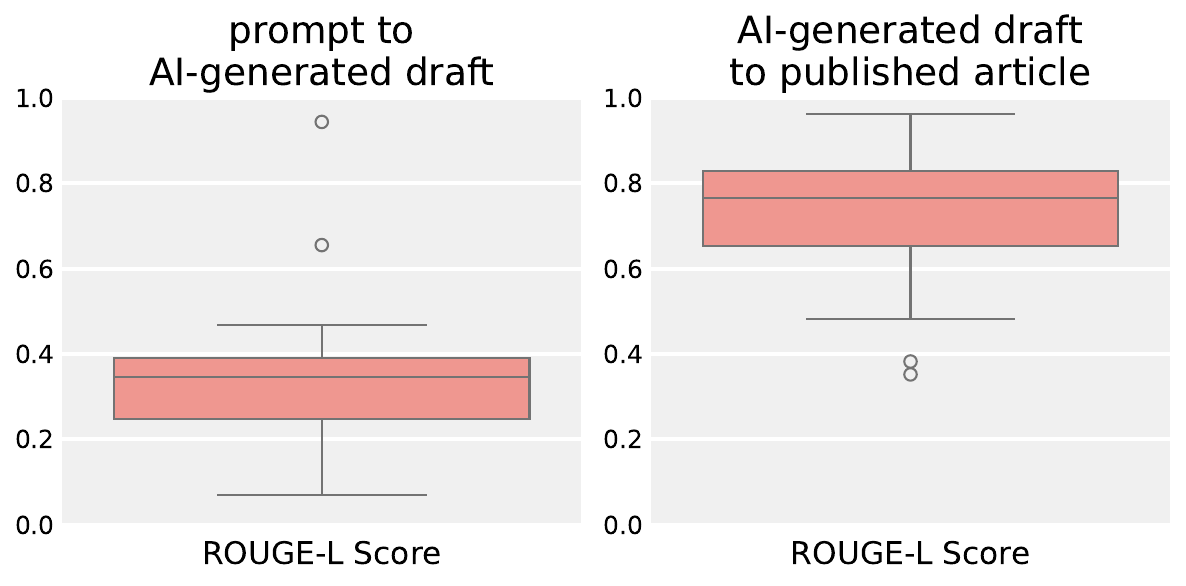}
    \footnotesize 
     \caption{Agency B}
    \end{subfigure}
    \caption{ROUGE-L scores for prompt to machine-generated output (left) and that output to published article text (right) for both agencies.} 
    \label{fig:rouge-app}
\end{figure}

\section{ROUGE-L}
\label{app:sec:rouge}
Figure~\ref{fig:rouge-app} depicts ROUGE-L distributions for prompts to GPT outputs and those outputs to published articles for each agency.